\documentclass{article}
\usepackage{spconf,amsmath,epsfig}
\usepackage{graphicx}
\usepackage{amsmath}
\usepackage{amssymb}
\usepackage{amsfonts}
\usepackage{booktabs}

\usepackage{framed,multirow}
\usepackage{latexsym}
\usepackage{verbatim}
\usepackage{color}
\usepackage{subfigure}

\let\OLDthebibliography\thebibliography
\renewcommand\thebibliography[1]{
  \OLDthebibliography{#1}
  \setlength{\parskip}{0pt}
  \setlength{\itemsep}{0pt plus 0.3ex}
}

\begin{document}\sloppy

\def\x{{\mathbf x}}
\def\L{{\cal L}}

\title{Uncertainty-aware U-Net for Medical Landmark Detection}
%
\name{Ziyang Ye, Haiyang Yu, and Bin Li}
\address{\{yezy20, hyyu20, libin\}@fudan.edu.cn\\
Shanghai Key Laboratory of Intelligent Information Processing\\
School of Computer Science, Fudan University}

\maketitle

\begin{abstract}
Heatmap-based methods play an important role in anatomical landmark detection. However, most current heatmap-based methods assume that the distributions of all landmarks are the same and the distribution of each landmark is isotropic, which may not be in line with reality. For example, the landmark on the jaw is more likely to be located along the edge and less likely to be located inside or outside the jaw. Manually annotating tends to follow similar rules, resulting in an anisotropic distribution for annotated landmarks, which represents the uncertainty in the annotation. To estimate the uncertainty, we propose a module named Pyramid Covariance Predictor to predict the covariance matrices of the target Gaussian distributions, which determine the distributions of landmarks and represent the uncertainty of landmark annotation. Specifically, the Pyramid Covariance Predictor utilizes the pyramid features extracted by the encoder of the backbone U-Net and predicts the Cholesky decomposition of the covariance matrix of the landmark location distribution. Experimental results show that the proposed Pyramid Covariance Predictor can accurately predict the distributions and improve the performance of anatomical landmark detection.
\end{abstract}

\begin{keywords}
uncertainty estimation, anatomical landmark detection, heatmap regression
\end{keywords}

\section{Introduction}
\label{sec:Introduction}

Anatomical landmark detection plays an important role in medical image analysis. Due to the advantage of likelihood modeling, researchers often utilize a heatmap to obtain the location of landmarks rather than directly regressing their coordinates \cite{kumar2020luvli}. The heatmap is a probability distribution of landmark location, and as a common practice, researchers usually use a Gaussian likelihood to estimate the distribution. 

\begin{figure}[!t]
\centering
 \subfigure[]{\includegraphics[width=3cm]{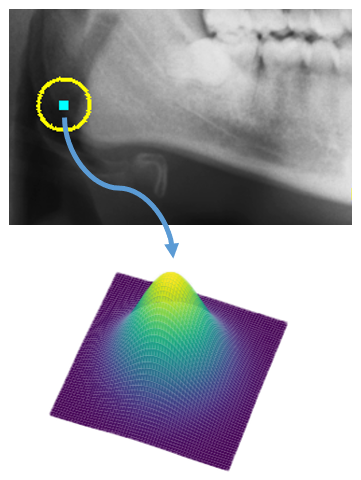}}
 \subfigure[]{\includegraphics[width=3cm]{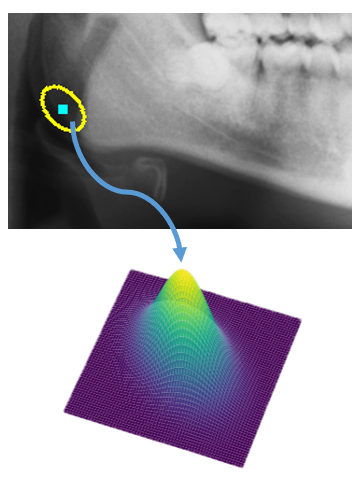}}
\caption{The left image shows the Gaussian distribution of one landmark used in most methods. However, the landmark is more likely to be located along the jaw and less likely to be located out in space or within the jaw, so the distribution illustrated in the right image is likely to be more representative.}
\label{fig:fig0}
\end{figure}

However, most current heatmap-based methods need predetermined heatmaps with fixed Gaussian kernels as prior knowledge, and this potentially introduces two assumptions into their approaches: 1) The distributions of all landmarks are the same. 2) The distribution of each landmark is isotropic. These hypotheses may not be consistent with reality. Firstly, since there may be differences in the difficulty of locating various landmarks, the distributions of landmarks may be different from each other. For landmarks that are easy to locate, such as those on clear edges, the variances of their distributions may be smaller, and vice versa. Secondly, for a single landmark, its distribution in different directions may also vary. For landmarks that lie on edges, their distributions may be more likely to be along the edge, thus introducing anisotropy \cite{payer2020uncertainty}. Fig.~\ref{fig:fig0} shows an example.


The aforementioned phenomenon shows the \textit{aleatoric} uncertainty of landmark detection. \textit{Aleatoric} uncertainty represents the noise inherent in the observations and is mainly caused by annotation ambiguities. It is difficult for humans to annotate anatomical landmarks with perfect accuracy. When annotating the same image several times, annotators may give different annotations, and the possibility can be described as a distribution. As mentioned before, different landmarks may have different distributions which are possibly anisotropic. However, previous methods simply utilize a fixed Gaussian kernel to get the heatmap, thus failing to model real distributions of landmarks. Specifically, these methods use a predetermined heatmap with the same variance and optimize the model by minimizing the loss between the predicted heatmap and the predetermined heatmap. Such predetermined heatmaps are not representative enough of the real annotation distributions, which may limit the performance of heatmap-based methods.

To this end, we propose a U-Net-based method to jointly predict the landmarks and their probability distributions. We use a Gaussian distribution to describe the possibility, and let the ground-truth landmark coordinate be the mean of the distribution. To predict the covariance matrix of the target distribution, we develop a Pyramid Covariance Predictor branch (shown in Fig.~\ref{fig:fig1}). Considering differences between low-level and high-level features, the branch processes the two kinds of features separately and fuses them to predict the covariance matrix. The experimental results show that the proposed Pyramid Covariance Predictor can indeed estimate the uncertainty of landmarks and improve the performance of anatomical landmark detection. Our contributions are as follows:

\begin{enumerate}
    \item We propose a method for uncertainty-aware landmark detection that jointly predicts the landmark and the covariance matrix of the target distribution.
    \item We design a module called Pyramid Covariance Predictor to discover the covariance matrix of the heatmap distribution. Visualizations demonstrate that the uncertainty our method estimates is more reasonable than that used in other methods.
    \item The experimental results show that our method achieves overall better performance compared with other methods.
\end{enumerate}

\section{Related Work}
\label{sec:Related Work}

\subsection{Medical Landmark Detection}

Early methods for medical landmark detection were based on a simple classifier, such as Random Forest \cite{ibragimov2015computerized, lindner2015fully}. Recently, deep-learning-based methods \cite{noothout2020deep,zhong2019attention,chen2019cephalometric} have been proven to be effective in medical landmark detection. Existing methods can be classified into three categories \cite{li2020structured}: 1) the coordinate-based approach \cite{noothout2020deep} that directly predicts the location of the landmarks by regression, 2) the graph-based approach \cite{li2020structured} that uses a graph to represent the structure of the landmarks, and 3) the heatmap-based approach.

Heatmap-based methods are the most common. These methods try to obtain the likelihood heatmap for each landmark and use this heatmap (sometimes with offset maps) to predict the locations of landmarks. For instance, in \cite{zhong2019attention}, a two-stage U-Net framework with attention mechanism and heatmap regression is developed to detect landmarks. In \cite{chen2019cephalometric}, an attentive feature pyramid fusion module is proposed to shape enhanced fusion features to improve accuracy. These methods generate the ground-truth heatmap using a symmetric Gaussian distribution with a fixed Gaussian kernel and optimize the model by minimizing the distance between predicted heatmaps and target heatmaps.

\subsection{Uncertainty Estimation}

There are two types of methods in uncertainty estimation: sampling-based and sampling-free. Sampling-based methods \cite{shridhar2019comprehensive,gal2016dropout,ayhan2018test} utilize the existing models and estimate the uncertainty by multiple evaluations. Popular methods, such as Bayesian neural networks \cite{shridhar2019comprehensive} and Monte Carlo dropout \cite{gal2016dropout}, all rely on several predictions by an ensemble of multiple networks or a dropout layer. Running the model on different augmentations\cite{ayhan2018test} is also an effective way.


Sampling-free methods are modified from previous architectures to compute uncertainty using a single network and the same dataset. For example, in \cite{ovadia2019can}, the authors introduced stochastic variational inference and temperature scaling into the model to evaluate uncertainty. In \cite{sensoy2018evidential}, the uncertainty is estimated by fitting a Dirichlet distribution. In other fields, researchers have also tried to introduce uncertainty into their methods to improve performance, such as face alignment \cite{chen2019face} and body pose estimation \cite{gundavarapu2019structured}.

\begin{figure}[!t]
\centering
\includegraphics[height=5.5cm]{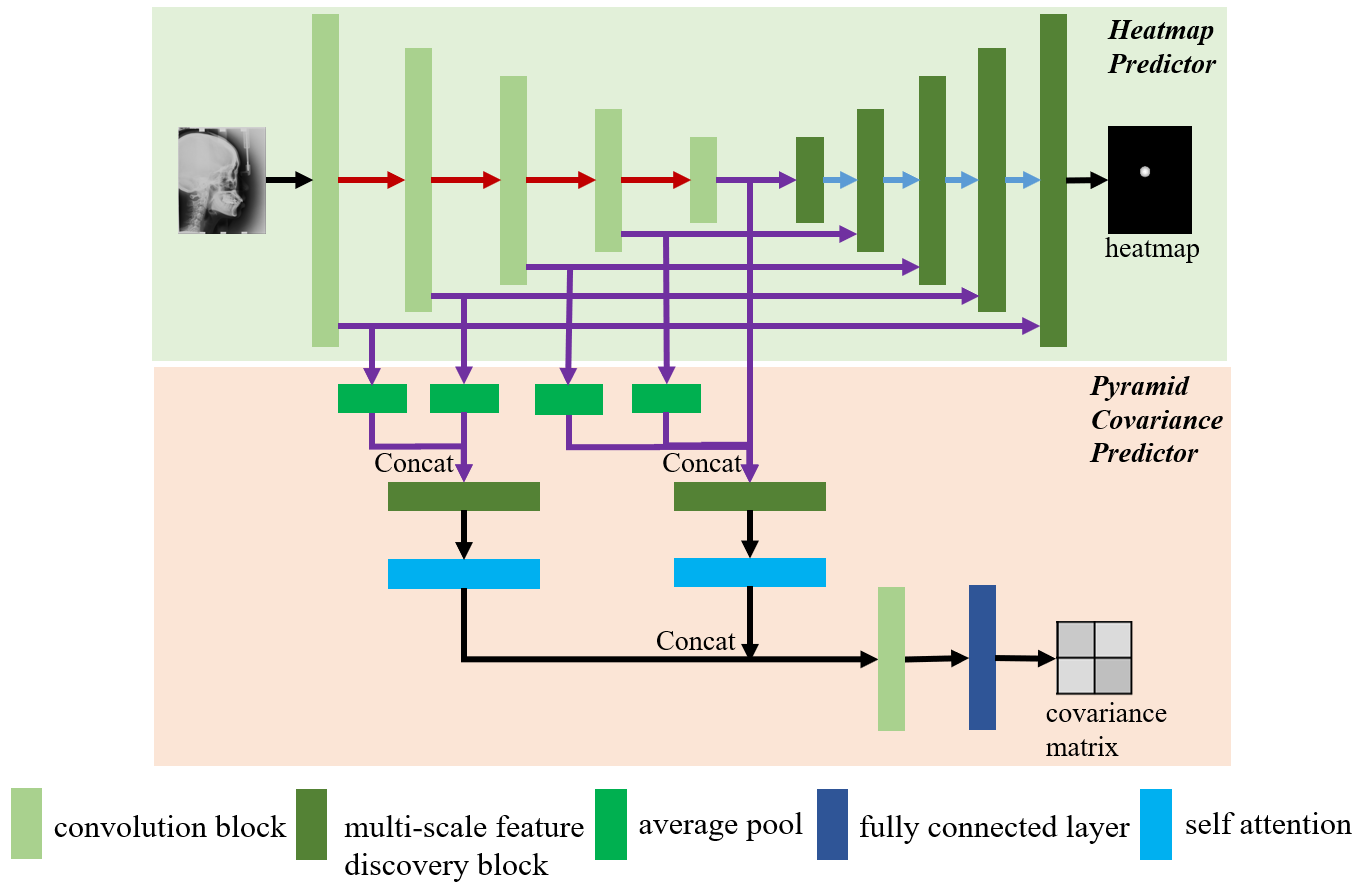}
\caption{Overview of our method. We add a branch to the U-Net architecture and fuse the low-level and high-level features separately. The low-level features are extracted by the first two layers of the encoder, and the high-level features are the output of remaining three layers.}
\label{fig:fig1}
\end{figure}

\section{Methods}
\label{sec:Method}

As shown in Fig.~\ref{fig:fig1}, we use a U-Net architecture as the heatmap predictor to obtain the heatmaps of landmarks. In this paper, we introduce a new branch, called Pyramid Covariance Predictor, to calculate the covariance matrix of the target distribution and estimate the annotating uncertainty. The details of our method are illustrated as follows.

\subsection{Heatmap Predictor}\label{section:HP}

\begin{figure}[!t]
\centering
\includegraphics[height=3cm]{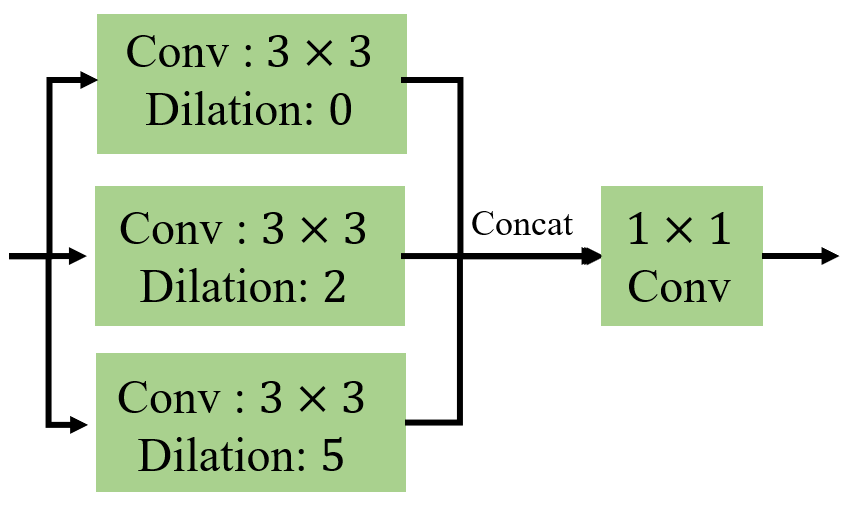}
\caption{The structure of the MSFD block consists of three convolution layers, with the concatenated feature as input. The outputs are then concatenated and processed by a $1 \times 1$ convolution to reduce the channel.}
\label{fig:fig2}
\end{figure}

We adopt the standard U-Net as the heatmap predictor due to its effectiveness and relatively simple structure. The standard U-Net architecture consists of an encoder and a decoder, where the encoder extracts the multi-scale features of input images and the decoder takes them as input to predict the heatmap. In our method, a VGG19 network \cite{simonyan2014very} pretrained on ImageNet Dataset \cite{krizhevsky2012imagenet} is adopted as the encoder. The decoder consists of consecutive modules. Each module contains an upsampling layer and a multi-scale feature discovery (MSFD) block (note that we omit the upsampling layer in Fig.~\ref{fig:fig1}). The upsampling layer upsamples the output from the last module using bilinear interpolation and combines the upsampled features with the image features from skip connections. The structure of the MSFD block is similar to the context-aware pyramid feature extraction (CPFE) module \cite{zhao2019pyramid}, which aims to fuse the combined features by the following two steps (as shown in Fig.~\ref{fig:fig2}): 1) extract multi-scale features by multiple atrous convolution layers with different dilation rates, and 2) concatenate them and reduce the channel. 

Finally, for $ N $ landmarks, the heatmap predictor outputs the heatmap $ \boldsymbol{\hat{H_i}} $, corresponding to the $i$-th landmark. For further processing, we calculate the predicted coordinate $ \boldsymbol{\hat{x}_i} $ from the heatmap. To utilize information from the whole heatmap and keep the gradient propagation, we use a weighted spatial mean to obtain $\boldsymbol{\hat{x}_i} $:

\begin{align}
 \label{eq:WSM}
 \boldsymbol{\hat{x}_i} = \frac{\Sigma_{\boldsymbol{x}}\delta\left(\boldsymbol{\hat{H_i}}\left(\boldsymbol{x}\right)\right)\boldsymbol{x}}{\Sigma_{\boldsymbol{x}}\delta\left(\boldsymbol{\hat{H_i}}\left(\boldsymbol{x}\right)\right)}
\end{align}
where $ \boldsymbol{\hat{H_i}}\left(\boldsymbol{x}\right) $ is the value at position $ \boldsymbol{x} $ in $ \boldsymbol{\hat{H_i}} $, $ \delta\left(\cdot\right) $ is the activation function.


\setlength{\tabcolsep}{3pt}
\begin{table*}[t]
\begin{center}
\caption{Comparison with other SOTA methods on IEEE ISBI Challenge 2015 Dataset.}
\label{table:table1}
\setlength{\tabcolsep}{2mm}{
\begin{tabular}{@{}ccccccccccc@{}}
\toprule

\multirow{3}{*}{Model} & \multicolumn{5}{c}{Validation Set} & \multicolumn{5}{c}{Test Set}  \\ \cmidrule(l){2-6} \cmidrule(l){7-11} & \multirow{2}{*}{MRE} & \multicolumn{4}{c}{SDR} & \multirow{2}{*}{MRE} & \multicolumn{4}{c}{SDR} \\ & & 2mm & 2.5mm & 3mm & 4mm & & 2mm & 2.5mm & 3mm & 4mm \\ \cmidrule(l){1-11}

Ibragimov \textit{et al.} \cite{ibragimov2015computerized} & 1.84 & 71.70 & 77.40 & 81.90  & 88.00 & - & 62.74 & 70.47 & 76.53 & 85.11 \\
Lindner \textit{et al.} \cite{lindner2015fully} & 1.67 & 74.95 & 80.28 & 84.56 & 89.68 & - & 66.11 & 72.00 & 77.63 & 87.42 \\
Arik \textit{et al.} \cite{arik2017fully} & - & 75.37 & 80.91 & 84.32 & 88.25 & - & 67.68 & 74.16 & 79.11 & 84.63 \\
Qian \textit{et al.} \cite{qian2019cephanet} & - & 82.50 & 86.20 & 89.30 & 90.60 & - & 72.40 & 76.15 & 79.65 & 85.90 \\
Chen \textit{et al.} \cite{chen2019cephalometric} & 1.17 & \textbf{86.67} & \textbf{92.67} & 95.54 & 98.53 & \textbf{1.48} & \textbf{75.05} & \textbf{82.84} & 88.53 & 95.05 \\
Lin \textit{et al.} \cite{lin2021structure} & 1.23 & 85.01 & 91.57 & 94.52 & 97.68 & 1.65 & 72.00 & 81.63 & 87.84 & 94.05 \\
Ours & \textbf{1.16} & 86.25 & 92.18 & \textbf{95.72} & \textbf{98.59} & \textbf{1.48} & 74.26 & 82.11 & \textbf{88.57} & \textbf{95.21} \\ \bottomrule
\end{tabular}}
\end{center}
\end{table*}
\setlength{\tabcolsep}{1.4pt}

\subsection{Pyramid Covariance Predictor}\label{section:CP}
The proposed Pyramid Covariance Predictor utilizes the pyramid features extracted by U-Net to predict the covariance matrix, which determines the distribution of the landmark location. For the multivariate location distribution, there are two approaches to building the target heatmap: Gaussian and Laplacian. Following \cite{payer2020uncertainty}, we typically choose the Gaussian function to represent the distributions of landmarks.

In previous works \cite{zhong2019attention,payer2016regressing}, the target heatmap $ \boldsymbol{H_i} $ of the $i$-th landmark can be described by an isotropic two-dimensional Gaussian function:

\begin{align}
 \label{eq:GF}
 \boldsymbol{H_i}\left(\boldsymbol{x};\sigma_i\right) = \frac{\gamma}{2\pi\sigma_i^2}\exp\left(-\frac{\|\boldsymbol{x} - \boldsymbol{x_i}\|^2_2}{2\sigma_i^2}\right)
\end{align}
where $ \boldsymbol{x_i} $ represents the location of the $i$-th landmark, $ \sigma_i $ is the standard deviation of the Gaussian distribution, and $ \gamma $ is a scaling factor that makes the function numerically stable. The mean of the distribution is set to the ground-truth landmark coordinate $ \boldsymbol{x_i} $ during training.


However, as aforementioned, isotropic Gaussian functions may be too simple to accurately represent the real distributions of landmarks. To better model the probability distributions, we use an anisotropic Gaussian function to estimate the uncertainty:

\begin{align}
 \label{eq:AGF}
 \boldsymbol{H_i}\left(\boldsymbol{x};\boldsymbol{\Sigma_i}\right) = \frac{\gamma \cdot exp\left(-\frac{1}{2}\left(\boldsymbol{x} - \boldsymbol{x_i}\right)^T\boldsymbol{\Sigma_i}^{-1}\left(\boldsymbol{x} - \boldsymbol{x_i}\right)\right)}{2\pi\sqrt{\left|\boldsymbol{\Sigma_i}\right|}}
\end{align}
Different from Eq.~\ref{eq:GF}, we use a full two-dimensional covariance matrix $ \boldsymbol{\Sigma_i} $ rather than a single value $ \sigma_i $ to represent the heatmap, thus introducing anisotropy into the distribution.

We follow \cite{kumar2020luvli} to predict the Cholesky decomposition of the covariance matrix in case of the illegal covariance matrix from direct regression. Details are further demonstrated in the Appendix. According to Cholesky Factorization Theorem \cite{schabauer2010toward}, the target covariance matrix can be decomposed as:


\begin{align}
 \label{eq:CD}
 \boldsymbol{\Sigma_i} = \boldsymbol{C_i}\boldsymbol{C_i}^T
\end{align}
where $ \boldsymbol{C_i} $ has the form of $ \begin{bmatrix} a_i & 0 \\ b_i & c_i \end{bmatrix} $. 

In our work, we calculate $ a_i $, $ b_i $, and $ c_i $ by utilizing the multi-scale features extracted by the encoder of U-Net architecture. The multi-scale features can be divided into two categories: low-level features and high-level features \cite{zhao2019pyramid}. Low-level features represent features that show the pattern of the image, such as color, texture, and edges, while high-level features contain more semantic information. Due to the difference between low-level features and high-level features, we process the two types of features separately. Following \cite{zhao2019pyramid}, we choose the features of the first two layers of the encoder as low-level features and choose the features of the remaining three layers as high-level features. For both low-level features and high-level features, we first apply average pooling to generate feature maps with the same resolution. Then, we concatenate them separately into two larger pyramid features. We fuse both features using the self-attention mechanism \cite{vaswani2017attention} since it has shown excellent ability in capturing long-range dependency. The self-attention mechanism is a Query-Key-Value architecture. It computes the scaled dot product of Query and Key and uses the result as a weight on Value. Formally, given an input feature $ X $, to compute its self-attention results, we first embed $ X $ to three different matrices $ \boldsymbol{Q} $, $ \boldsymbol{K} $, $ \boldsymbol{V} $, representing Query, Key, and Value:

\begin{align}
 \boldsymbol{Q} = \boldsymbol{XW_Q}, \quad\boldsymbol{K} = \boldsymbol{XW_K}, \quad\boldsymbol{V} = \boldsymbol{XW_V}
\end{align}
Then the self-attention results can be computed by:

\begin{align}\label{formula:sa}
 \boldsymbol{S} = \sigma\left(\frac{\boldsymbol{QK}^T}{\sqrt{d}}\right)\boldsymbol{V}
\end{align}
where $ d $ is the dimension of $\boldsymbol{K}$, $ \sigma $ is the softmax function.

Finally, we concatenate the output features and predict the Cholesky decomposition through a fully connected layer. Thus, the covariance matrix can be computed using its Cholesky decomposition (shown in Eq. \ref{eq:CD}).

\subsection{Loss Function}\label{section:LF}

The loss function of our method, similar to the negative log-likelihood of the anisotropic Gaussian function, can be expressed as follows:

\begin{align}\label{formula:loss}
 L = \left(\boldsymbol{\hat{x}_i} - \boldsymbol{x_i}\right)^T\boldsymbol{\hat{\Sigma}}_i^{-1}\left(\boldsymbol{\hat{x}_i} - \boldsymbol{x_i}\right) + \alpha log\left|\boldsymbol{\hat{\Sigma}_i}\right|
\end{align}
where $ \boldsymbol{\hat{x}_i} $ and $ \boldsymbol{x_i} $ are the predicted and ground truth locations of the $ i $-th landmark, respectively. $ \boldsymbol{\hat{\Sigma}_i} $ is the predicted covariance matrix.

The first term in Eq.~\ref{formula:loss}  is the squared Mahalanobis distance between the predicted landmark and the ground truth. The second term in Eq.~\ref{formula:loss} is a regularization to prevent the distribution from becoming excessively flattened. $ \alpha $ is a hyperparameter to adjust the weight of the regularization term. In our experiment, we empirically set it to 0.1. 

\setlength{\tabcolsep}{3pt}
\begin{table*}[t]
\begin{center}
\caption{Results of ablation experiments.}
\label{table:table2}
\begin{tabular}{@{}ccccccccccc@{}}
\toprule
\multirow{3}{*}{Model} & \multicolumn{5}{c}{Validation Set} & \multicolumn{5}{c}{Test Set}  \\ \cmidrule(l){2-6} \cmidrule(l){7-11} & \multirow{2}{*}{MRE} & \multicolumn{4}{c}{SDR} & \multirow{2}{*}{MRE} & \multicolumn{4}{c}{SDR} \\ & & 2mm & 2.5mm & 3mm & 4mm & & 2mm & 2.5mm & 3mm & 4mm \\ \cmidrule(l){1-11}
U-Net & 1.24 & 84.84 & 90.52 & 93.75 & 97.40 & 1.61 & 71.89 & 80.63 & 86.36 & 93.68 \\
Exp-U-Net & 1.23 & 84.59 & 91.64 & 94.87 & 98.28 & \textbf{1.48} & \textbf{75.05} & \textbf{82.84} & \textbf{88.68} & 94.42 \\
Ours & \textbf{1.16} & \textbf{86.25} & \textbf{92.18} & \textbf{95.72} & \textbf{98.59} & \textbf{1.48} & 74.26 & 82.11 & 88.57 & \textbf{95.21} \\ \bottomrule
\end{tabular}
\end{center}
\end{table*}
\setlength{\tabcolsep}{1.4pt}

\begin{figure}[!t]
\centering
\includegraphics[height=7cm]{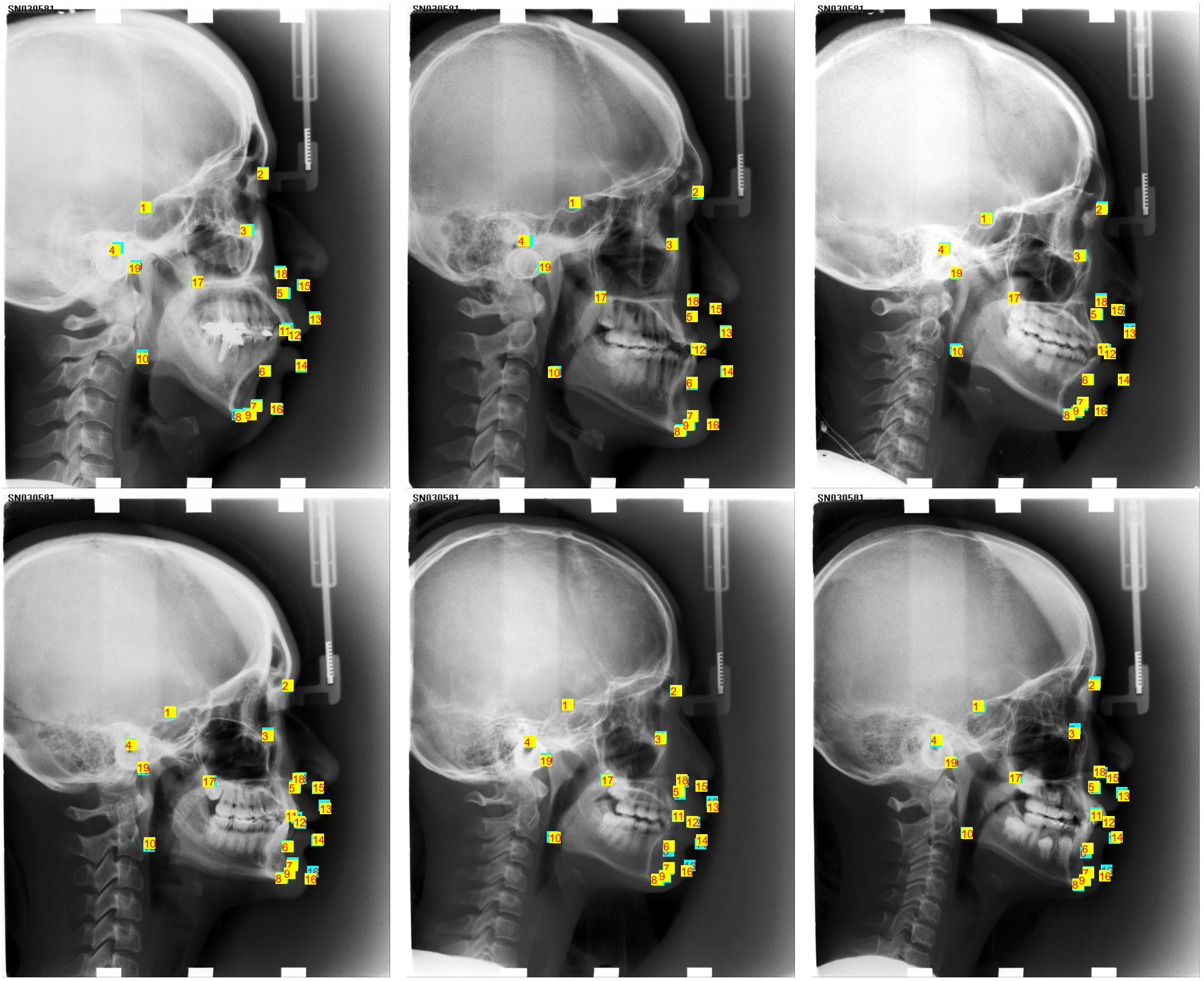}
\caption{The result samples of our method. The images in the first row are from the validation set, and the others are from the test set. The indigo and yellow points represent the ground truth landmarks and predicted landmarks, respectively. The red numbers indicate the indices of landmarks.}
\label{fig:fig3}
\end{figure}

\section{Experiments}
\label{sec:Experiments}

\subsection{Experimental Setting}\label{section:ES}

\textbf{Datasets.} We conduct experiments on IEEE ISBI Challenge 2015 Dataset \cite{wang2016benchmark}, which is widely used in cephalometric landmark detection tasks. It contains 400 cephalometric radiograph samples with a size of $ 1935 \times 2400 $, each labeled with 19 landmarks by two doctors. We choose the average of two annotations as the ground truth. Following previous methods \cite{chen2019cephalometric,oh2020deep}, we set the sizes of the training set, validation set, and test set to 150, 150, and 100, respectively. The result samples are illustrated in Fig.\ref{fig:fig3}. 

\noindent\textbf{Evaluation.} We use two metrics to evaluate the performance: Mean Radial Error (MRE) and Successful Detection Rate (SDR) in different radii (2mm, 2.5mm, 3mm, and 4mm). These standards are computed as follows:

\begin{gather}
 \text{MRE} = \frac{\sum_{i=1}^{n}\sqrt{\Delta x_i^2 + \Delta y_i^2}}{n} \\
 \text{SDR} = \frac{N_{acc}}{N_{all}} \times 100\%
\end{gather}
where $ \Delta x_i $ and $ \Delta y_i $ are the absolute differences between the $i$-th ground truth landmark and the corresponding predicted landmark in the x and y axis, respectively. $ N_{acc} $ is the number of successful detections and $ N_{all} $ is the number of detections. A successful detection is defined as the real distance between the two landmarks being lower than the precision radius.

\noindent {\bf Implementation Details.}  For a fair comparison, we resize the cephalometric radiographs to $ 800 \times 640 $ and normalize them for further training. We use PyTorch to build our framework and use Adam \cite{kingma2014adam} as our optimizer. The learning rate is set to 3e-4, and the weight decay is 1e-4.

\subsection{Comparison with Other Methods}\label{section:CWOM}

We select six representative landmark detection methods for comparison. The experimental results shown in Tab. \ref{table:table1} demonstrate that our method achieves better performance on MRE and outperforms existing SOTA methods on SDR in 3mm and 4mm radii. For localization evaluated by SDR in 2mm and 2.5mm radii, our method also achieves comparable results with the SOTA method \cite{chen2019cephalometric}. 

The results show that our method has an overall better result, but compared with Chen \textit{et al.} \cite{chen2019cephalometric}, our method is slightly less effective in terms of more accurate predictions. The method in \cite{chen2019cephalometric} uses not only a heatmap but also two offset maps to predict the landmarks. In their approach, the heatmap only provides a rough prediction, and the offset maps are utilized to precisely locate the position. However, due to the proposed Pyramid Covariance Predictor, our method can improve the performance of the heatmap and achieve comparable results using only heatmaps. Additionally, our method has a better result in larger precision radii. More comparison experiments are shown in the Appendix.

\subsection{Ablation Studies}\label{section:AS}

In this section, we conduct ablation studies on the Pyramid Covariance Predictor and use the classic U-Net framework as the baseline model. To demonstrate the effectiveness of the proposed Pyramid Covariance Predictor, we also compare our model with the traditional U-Net framework with similar computation and parameter levels (Exp-U-Net). Exp-U-Net is constructed by adding convolution layers.

The experimental results are shown in Tab.~\ref{table:table2}. Our method outperforms the compared methods in all metrics on the validation set. For the test set, our method also achieves better results in MRE and SDR on 4mm radius. Although Exp-U-Net has better performance on the test set in other precision radii, the performance gap between our method and Exp-U-Net gradually narrows with an increase in the precision radius. These results also show that our method can perform better in larger precision radii.

\begin{figure}[!t]
\centering
\includegraphics[height=7cm]{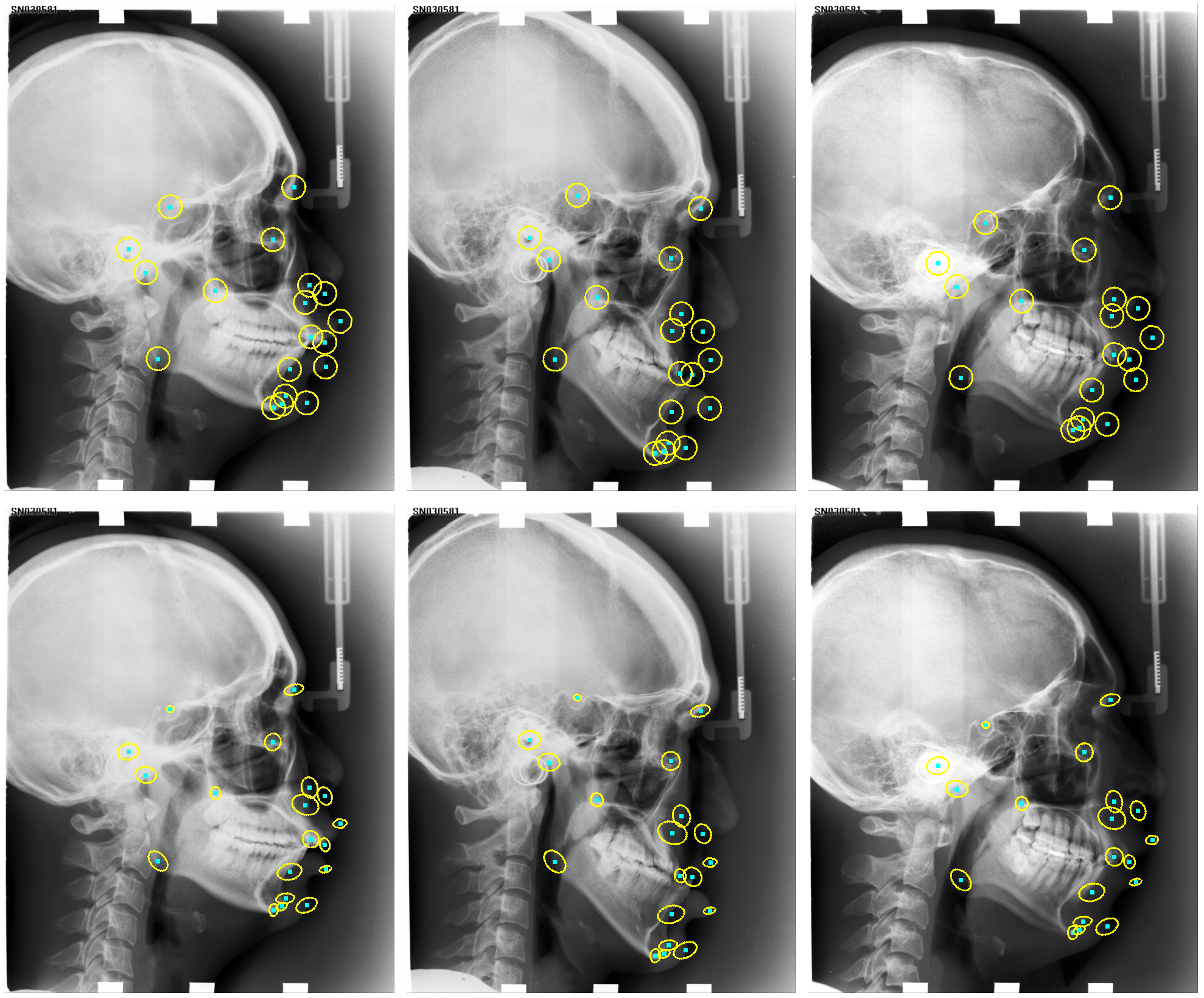}
\caption{A comparison between our predicted Gaussian distributions (the lower three images) and the predetermined Gaussian distributions generated by Chen \textit{et al.} \cite{chen2019cephalometric} (the upper three images) is shown. Cyan points represent the landmarks, and yellow ellipses indicate the distribution of each landmark.}
\label{fig:fig4}
\end{figure}

\subsection{Uncertainty Estimation}\label{section:UE}

In our work, we also predict the covariance matrix and obtain the distribution for each landmark. The distributions we predict differ from the predetermined distributions used in most of the current heatmap-based methods, as illustrated in Fig.~\ref{fig:fig4}. We also measure the uncertainty of the predicted landmarks and compare it with the predicted uncertainty; the details are discussed in the Appendix.

The predicted distributions of our method are intuitively in line with the annotation uncertainty. For each ellipse in the right image of Fig.~\ref{fig:fig4}, the direction of the major axis conforms to the direction of the edge to a certain degree. Therefore, the possibility along the normal direction decreases more rapidly than that along the edge, which is in accordance with our cognition. The results show that a heatmap-based landmark detection framework can also be used to model the uncertainty.

In addition, some neighbor ellipses overlap in the left image of Fig.~\ref{fig:fig4}, especially in the areas where landmarks are relatively densely distributed. This indicates that those pixels in the overlapping area may be detected as multiple landmarks with similar probability, which may confuse the model. However, in the generated distributions of our method, each distribution area can be well separated, which further demonstrates the effectiveness of our method.

\section{Conclusion}
\label{sec:Conclusion}

In this paper, we propose the Pyramid Covariance Predictor to discover the \textit{aleatoric} uncertainty of each landmark. The Pyramid Covariance Predictor makes use of multi-scale features and predicts the covariance matrix of the anisotropic Gaussian distribution to better represent the distributions of landmarks. Experimental results show that our method can estimate the uncertainty to some degree and achieve overall better results than other compared methods in cephalometric landmark detection.


\bibliographystyle{IEEEbib}
\bibliography{icme2023template}


\newpage
\noindent \Large \textbf{Appendix}
\normalsize
\renewcommand\thesection{\Roman{section}}
\setcounter{section}{0}
\section{Further Demonstration of Cholesky Decomposition}
\label{sec:sec1}

In the paper, the Pyramid Covariance Predictor predicts the Cholesky decomposition of the covariance matrix rather than the covariance itself. The reason we do not directly predict the covariance is that there is a relationship among elements of the covariance matrix. Considering that the covariance matrix is a positive semi-definite matrix, its determinant is non-negative:

\begin{align}
 \label{eq:DET}
 det\left(\boldsymbol{\Sigma_i}\right) = cov\left(\boldsymbol{x}, \boldsymbol{x}\right)cov\left(\boldsymbol{y}, \boldsymbol{y}\right) - cov\left(\boldsymbol{x}, \boldsymbol{y}\right)^2 \geq 0
\end{align}

Therefore, if we directly predict the elements of the matrix by regression, the outputs may be illegal since the outputs of the fully connected layer may not follow Eq.~\ref{eq:DET}. However, the elements of its Cholesky decomposition are independent of each other. So we choose to predict these values by regression.

Also, when using Cholesky decomposition, we hypothesize that the target covariance matrix is a positive definite matrix. This assumption is reasonable since in Eq.~\ref{eq:DET}, $ det\left(\boldsymbol{\Sigma_i}\right) = 0 $ if and only if the correlation coefficient of $ \boldsymbol{X} $ and $ \boldsymbol{Y} $ equals 1 or -1, which is nearly impossible in the reality. Regarding the covariance matrix as a positive definite matrix can simplify the problem.

It has been proven that the Cholesky decomposition of a positive definite matrix always exists. Let $\boldsymbol{A} \in \mathbb{R}^{n \times n}$ be positive definite. Obviously, for $n = 1$, you can take the square root as the Cholesky decomposition. Assuming that its Cholesky decomposition exists for $\boldsymbol{A} \in \mathbb{R}^{(n-1) \times (n-1)}$, then for $\boldsymbol{A} \in \mathbb{R}^{n \times n}$, it can be partitioned as:

\begin{align}
 \boldsymbol{A} = \begin{bmatrix} \boldsymbol{\hat{A}} & \alpha \\ \alpha^T & \beta \end{bmatrix}
\end{align}
where $ \boldsymbol{\hat{A}} \in \mathbb{R}^{(n-1) \times (n-1)} $.

Obviously $ \boldsymbol{\hat{A}} $ is positive definite, so it has a Cholesky decomposition $ \boldsymbol{\hat{A}} = \boldsymbol{\hat{L}}\boldsymbol{\hat{L}^T} $. Let:

\begin{gather}
 \label{eq:proof}
 \boldsymbol{L_1} = \begin{bmatrix} \boldsymbol{\hat{L}} & 0 \\ 0 & 1 \end{bmatrix} \\
 \boldsymbol{L_2} = \begin{bmatrix} \boldsymbol{I} & 0 \\ \gamma^T & 1 \end{bmatrix} \\
 \boldsymbol{L_3} = \begin{bmatrix} \boldsymbol{I} & 0 \\ 0 & \lambda \end{bmatrix} \\
 \boldsymbol{L} = \boldsymbol{L_1}\boldsymbol{L_2}\boldsymbol{L_3} = \begin{bmatrix} \boldsymbol{\hat{L}} & 0 \\ \gamma^T & \lambda \end{bmatrix}
\end{gather}
where $ \gamma = \boldsymbol{\hat{L}^{-1}}\alpha $, $ \lambda^2 = \beta - \alpha^T\boldsymbol{A}^{-1}\alpha $. Then we have:

\begin{align}
 \boldsymbol{A} = \boldsymbol{L}\boldsymbol{L^T}
\end{align}
where $ \boldsymbol{L} $ is the Cholesky decomposition of $ \boldsymbol{A} $.

Note that the Cholesky decomposition is not always unique since the sign of $ \lambda $ is not fixed. However, with all-positive elements, the Cholesky decomposition is unique. In our method, we force all elements of the predicted Cholesky decomposition to be positive for convenience by:

\begin{align}
 f(x) = \begin{cases}
 x+1, & if \quad x \geq 0 \\
 e^x, & otherwise
 \end{cases}
\end{align}

\setlength{\tabcolsep}{3pt}
\begin{table}[t]
\begin{center}
\caption{Results of comparison experiments on Digital Hand Atlas Dataset}
\label{table1}
\begin{tabular}{@{}ccccc@{}}
\toprule
\multirow{2}{*}{Model} & \multirow{2}{*}{MRE} & \multicolumn{3}{c}{SDR} \\ & & 2mm & 4mm & 10mm \\ \cmidrule(l){1-5}
Lindner et al\cite{lindner2014robust} & 0.85 & 93.67 & 98.95 & 99.94 \\
Štern et al\cite{vstern2016local} & 0.80 & 92.20 & 98.45 & 99.95 \\
Urschler et al\cite{urschler2018integrating} & 0.80 & 92.19 & 98.46 & 99.95 \\
Payer et al\cite{payer2019integrating} & 0.66 & 94.99 & 99.27 & \textbf{99.99} \\
Kang et al\cite{kang2021accurate} & \textbf{0.64} & 96.04 & \textbf{99.66} & 99.98 \\
Ours & \textbf{0.64} & \textbf{96.95} & 99.52 & \textbf{99.99} \\ \bottomrule
\end{tabular}
\end{center}
\end{table}
\setlength{\tabcolsep}{1.4pt}

\setlength{\tabcolsep}{3pt}
\begin{table*}[t]
\begin{center}
\caption{Comparison with or without our covariance predictor on IEEE ISBI Challenge 2015 Dataset.}
\label{table2}
\setlength{\tabcolsep}{2mm}{
\begin{tabular}{@{}ccccccccccc@{}}
\toprule
\multirow{3}{*}{Model} & \multicolumn{5}{c}{Validation Set} & \multicolumn{5}{c}{Test Set}  \\ \cmidrule(l){2-6} \cmidrule(l){7-11} & \multirow{2}{*}{MRE} & \multicolumn{4}{c}{SDR} & \multirow{2}{*}{MRE} & \multicolumn{4}{c}{SDR} \\ & & 2mm & 2.5mm & 3mm & 4mm & & 2mm & 2.5mm & 3mm & 4mm \\ \cmidrule(l){1-11}
Chen et al\cite{chen2019cephalometric} & 1.17 & \textbf{86.67} & \textbf{92.67} & \textbf{95.54} & \textbf{98.53} & 1.48 & 75.05 & 82.84 & 88.53 & 95.05 \\
Chen et al + covariance predictor & \textbf{1.15} & 86.60 & 92.21 & 95.50 & 98.31 & \textbf{1.39} & \textbf{76.00} & \textbf{83.32} & \textbf{89.74} & \textbf{96.32} \\ \bottomrule
\end{tabular}}
\end{center}
\end{table*}
\setlength{\tabcolsep}{1.4pt}

\section{More Results of Comparison Experiments}
\label{sec:sec2}
\begin{figure*}[!t]
\centering
\includegraphics[height=7cm]{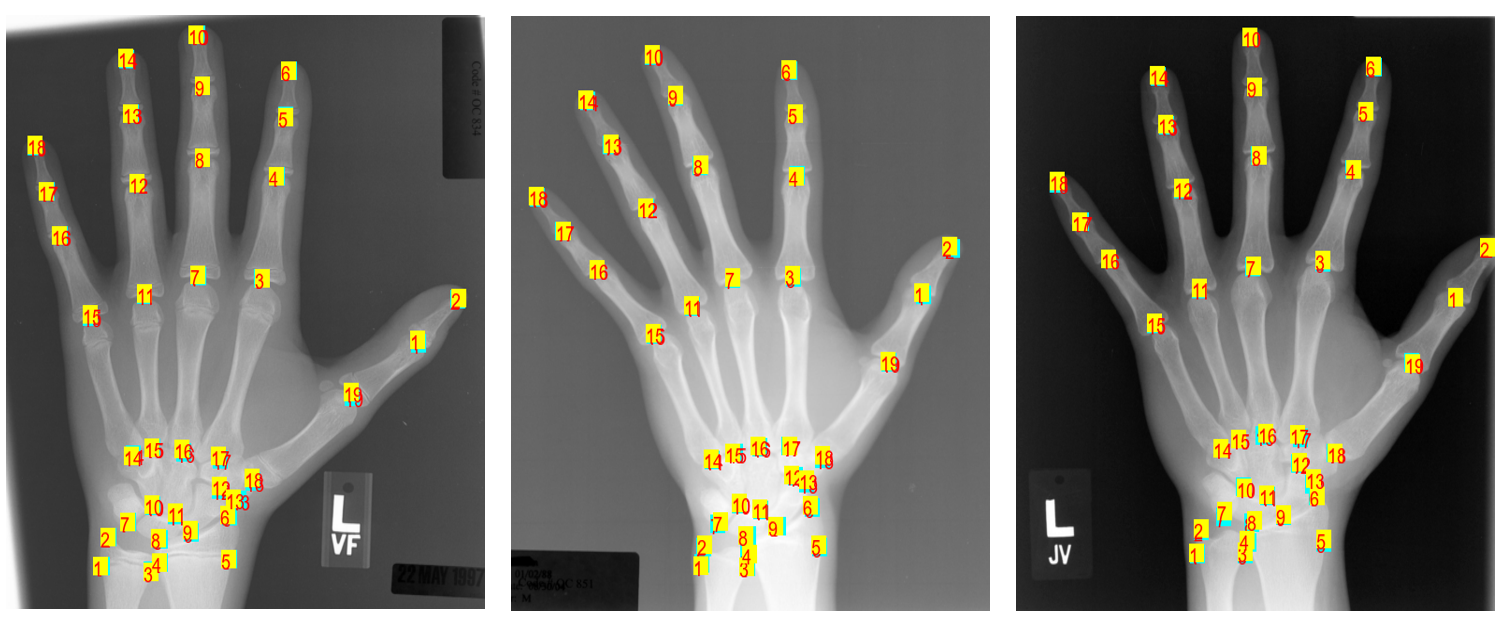}
\caption{The result samples of our method on Digital Hand Atlas Dataset.}
\label{fig0}
\end{figure*}

We conduct more comparison experiments on Digital Hand Atlas Dataset \cite{gertych2007bone}. Digital Hand Atlas Dataset consists of 895 X-ray images
of left hands with an average resolution of $ 1563 \times 2169 $ and 37 landmarks annotated in each image. Following \cite{payer2019integrating,kang2021accurate}, we use the same three-fold cross-validation, where images are divided into approximately 600 training and 300 testing images per fold. We evaluate the performance using MRE and SDR in 2mm, 4mm, and 10mm. The result samples are shown in Fig.~\ref{fig0}.

We select five representative landmark detection methods for comparison. The experimental results shown in Table \ref{table1} demonstrate that our method achieves better performance on MRE and outperforms existing SOTA methods on SDR in 2mm and 10mm radii. For localization evaluated by SDR in 4mm radius, our method also has a near SOTA performance. The extended experiment shows the generalization ability of our method. Due to the effectiveness of uncertainty estimation, our method can not only be used in cephalometric landmark detection, but can also be extended to other medical landmark detection tasks.

We also conduct additional experiments by combining the proposed Pyramid Covariance Predictor with the method of Chen et al \cite{chen2019cephalometric}, and the results are shown in Table \ref{table2}. Benefiting from the proposed Pyramid Covariance Predictor, the method \cite{chen2019cephalometric} achieves comparable performance on validation sets, and outperforms the baseline model on test sets by a clear margin. The results verify the effectiveness of the proposed Pyramid Covariance Predictor.

\section{Further Analysis on Uncertainty Estimation}
\label{sec:sec3}

\begin{figure}[!t]
\centering
\includegraphics[height=5.5cm]{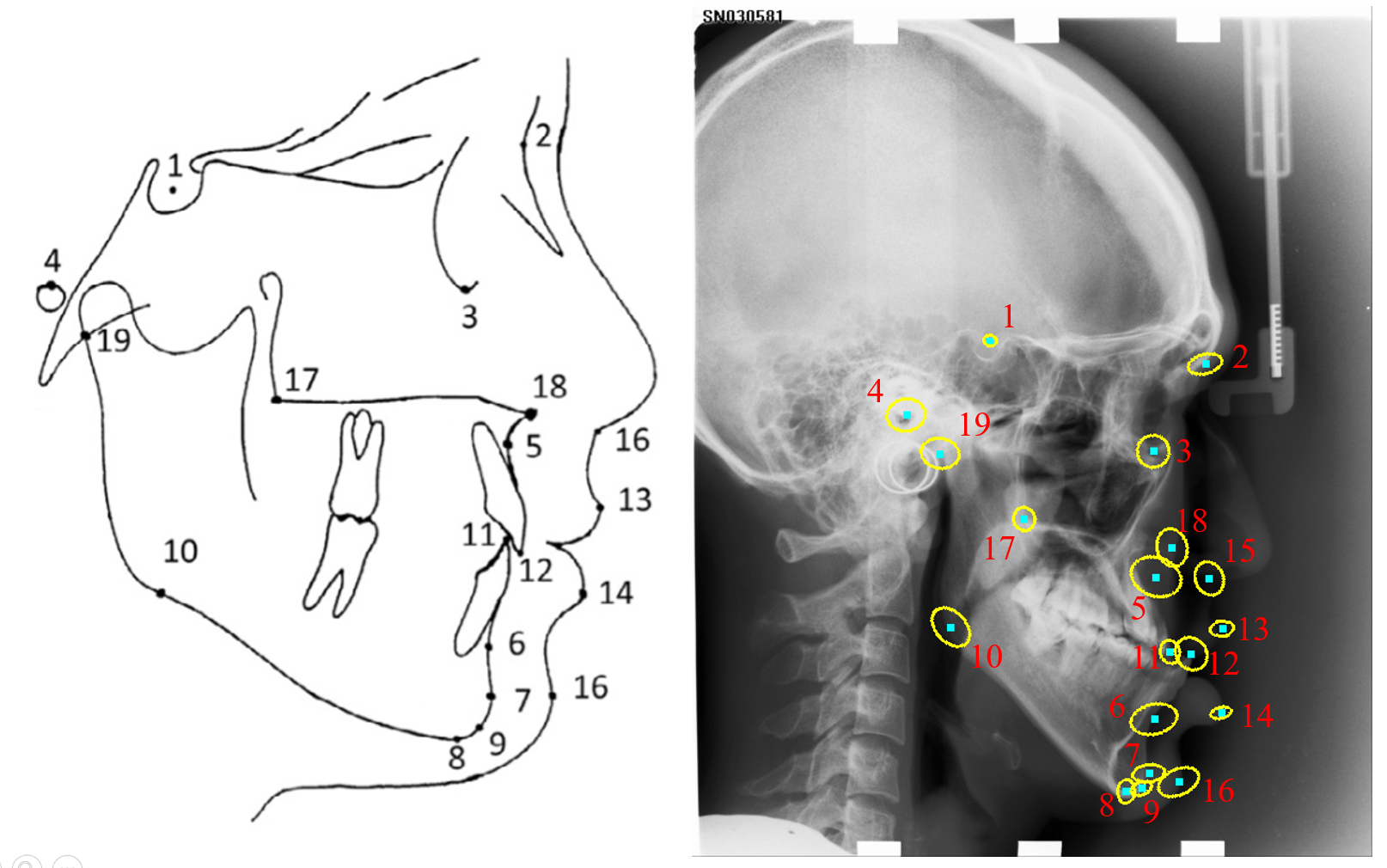}
\caption{The ordering of the cephalometric landmarks. The left image is the original image of cephalometric tracing \cite{wang2016benchmark}, and the right image is the index label on our illustrated image.}
\label{fig1}
\end{figure}

In our paper, we figure out that the Gaussian distributions of the landmarks should be anisotropic and unlikely to overlap. The landmark predictions of our method reflect that. For reference, we show the ordering of the cephalometric landmarks in Fig.~\ref{fig1}.

\begin{figure}[!t]
\centering
\includegraphics[height=8cm]{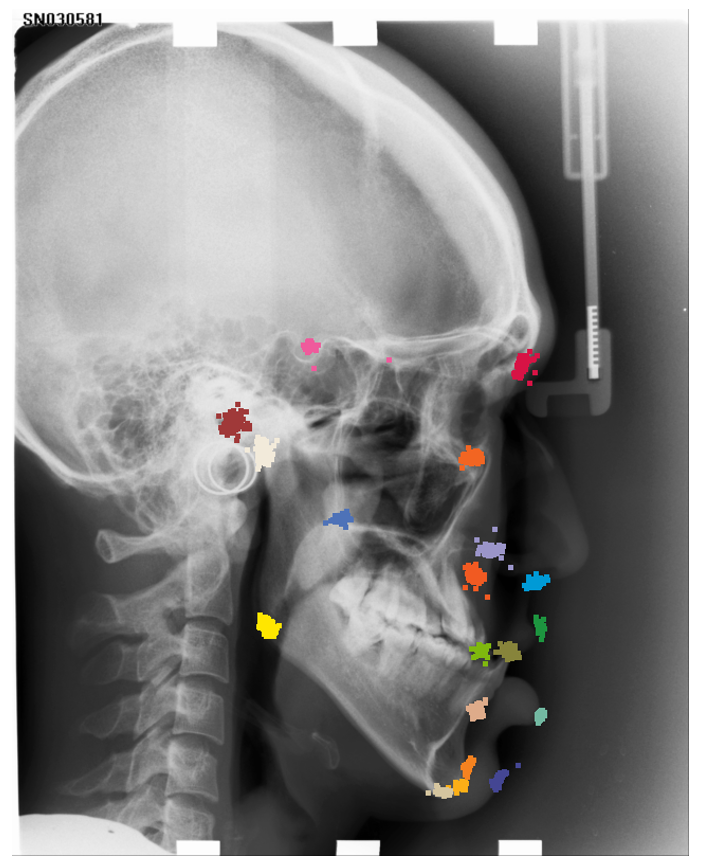}
\caption{Sample image of all the displacement landmarks.}
\label{fig2}
\end{figure}

We measure the uncertainty of the predicted landmarks. Firstly, for every landmark, we compute all displacement vectors between predicted locations and the ground truth locations of all images in the validation set and test set. Then we show them in one image. For landmarks on that image, we obtain the coordinates of displacement landmarks by adding the ground truth locations and every corresponding displacement vector separately. We show the example image in Fig.~\ref{fig2}, which demonstrates the uncertainty of every landmark. Consistent with what we think, the distributions of displacement landmarks are anisotropic, mainly along the edges. Also, different landmarks have different distribution ranges, which shows that the difficulties in predicting different landmarks are various. 

Note that there are few overlaps among the displacement landmarks. In most of the current heatmap-based methods, since they use a fixed heatmap for every landmark, the distribution ranges may overlap. For example, in our paper, we illustrate the predetermined Gaussian distributions generated by Chen et al \cite{chen2019cephalometric}, which is also utilized in many other heatmap-based methods. There are many overlaps in that image, for example, distributions of landmark 11 and landmark 12. However, the displacement landmarks of landmark 11 and 12 in Fig.~\ref{fig2} show that 
there are no overlaps between them, which indicates the disadvantages of a fixed heatmap.

However, the distributions we predict are not exactly consistent with the distributions of displacement landmarks. There are still some deviations between the directions of our predicted distributions and those of the displacement landmarks. For example, for landmark 2, the distribution of the displacement landmarks is along the nasal root, but our predicted distribution deflects from the nasal root. It indicates that our method can be further optimized and thus achieve better performance.

\end{document}